\documentclass{article}
\usepackage[utf8]{inputenc}

\usepackage{lipsum}
\usepackage{latexsym}
\usepackage{subfigure}
\usepackage[ruled,vlined]{algorithm2e}
\usepackage{amsmath}
\usepackage{algpseudocode}
\usepackage[table]{xcolor}
\usepackage{aaai21}  
\usepackage{times}  
\usepackage{helvet} 
\usepackage{courier}  
\usepackage[hyphens]{url}  
\usepackage{hyperref}
\usepackage[hyphenbreaks]{breakurl}
\usepackage[switch]{lineno}  
\usepackage{graphicx} 
\usepackage{natbib}  
\usepackage{caption} 
\title{What's the best place for an AI conference, Vancouver or \underline{~~~~~~}: \\Why completing comparative questions is difficult}
\author{Avishai Zagoury,\textsuperscript{\rm 1}
Einat Minkov,\textsuperscript{\rm 2}\thanks{Work done at Google Research.}
Idan Szpektor,\textsuperscript{\rm 1}
William W. Cohen\textsuperscript{\rm 1} \\
}
\affiliations {
    \textsuperscript{\rm 1} Google Research,
    \textsuperscript{\rm 2} University of Haifa \\
    avishaiz,szpektor,wcohen@google.com, einatm@is.haifa.ac.il \\
}
\date{August 2020}

\begin{document}


\maketitle

\begin{abstract}
Although large neural language models (LMs) like BERT can be finetuned to yield state-of-the-art results on many NLP tasks, it is often unclear what these models actually learn.  Here we study using such LMs to fill in entities in human-authored comparative questions, like ``Which country is older, India or \underline{~~~~~~}?''---i.e., we study the ability of neural LMs to \emph{ask} (not \emph{answer}) reasonable questions.  We show that accuracy in this fill-in-the-blank task is well-correlated with human judgements of whether a question is reasonable, and that these models can be trained to achieve nearly human-level performance in completing comparative questions in three different subdomains. However, analysis shows that what they learn fails to model any sort of broad notion of which entities are semantically comparable or similar---instead the trained models are very domain-specific, and performance is highly correlated with co-occurrences between specific entities observed in the training set.  This is true both for models that are pretrained on general text corpora, as well as models trained on a large corpus of comparison questions. Our study thus reinforces recent results on the difficulty of making claims about a deep model's world knowledge or linguistic competence based on performance on specific benchmark problems. We make our evaluation datasets publicly available to foster future research on complex understanding and reasoning in such models at standards of human interaction.  
\end{abstract}

\section{Introduction}

Neural language models like BERT~\cite{bert19} that are pretrained on very large amounts of unlabeled texts produce highly informative contextual word representations which have been proven to capture various linguistic phenomena, such as syntax and coreference~\cite{Tenney:2019,belinkovCL20}. Recently, it has been argued that these models also encode some factual and common-sense knowledge, presenting a potential alternative to curated knowledge bases for question answering tasks \cite{petroniEMNLP19,robertsARXIV20,jiangTACL2020}. Yet, it is an open question to what extent neural LMs accurately represent the meaning encapsulated in a natural language utterance, and whether they can truly support knowledge-based semantic reasoning with the factual knowledge that they encode within them~\cite{talmorARXIV19,richardsonAAAI2020,kassnerACL20}.

In this work, we introduce and explore a new task which requires world knowledge, language understanding and semantic reasoning, namely, learning to generate {meaningful} comparative questions. Consider, for example, the question, ``\emph{is the area of Montana larger than Arizona?}'' this question is sensible and non-obvious, as Montana and Arizona are both large states. Likewise, comparing between the weights of rhinos and hippopotamuses  seems natural, as both are large animals. In contrast, asking about the relative height of a lion and a mouse is perceived as nonsensical.

There are several motivations for studying the task of automatically generating sensible comparative questions. 
In the context of human-machine interaction, comparative questions serve as interest-invoking elements in conversational search and exploration dialogues \cite{tsurelWSDM17,Szpektor:2020}. They call for active participation of the user, and can serve as a rhetorical method for shifting the focus of the conversation, thus keeping it dynamic and evolving.
From a linguistic perspective, generating meaningful comparisons combines high-level reasoning with factual knowledge as well as modeling the likely knowledge of others. The probing of state-of-the-art language model on this task can indicate how far these models go in terms of ``understanding language'' versus ``memorizing facts''.

In this paper, we evaluate state-of-the-art neural language models for automatically generating meaningful comparisons.  To make evaluation less subjective we frame this task as as a fill-the-blank task, where we mask a single entity mention in an actual comparative question, and ask the model to complete the question. Unlike many prior studies, we evaluate the generality and robustness of what is learned by partitioning the task by subdomain.  In particular, we consider comparative questions posted by human users on question answering websites about {\it animals}, {\it cities and countries}, and {\it NBA players}. As illustrated in Table~\ref{tab:examples}, these questions compare entities along various domain-related properties that are both factual (e.g., ``bigger'') and subjective (``more loyal''). Automatically completing such questions requires a combination of linguistic understanding, world knowledge, and common-sense reasoning. We make our test-sets publicly available, to be used as a benchmark and to encourage further research on this task.

\begin{table}[t]
\small
\begin{tabular}{p{0.46\textwidth}}
\hline
(1) What is bigger, a MASK or a cougar? $<${\it great dane (Q5414)}$>$ \\
(2) What is more loyal a MASK or pigeon? $<${\it parrot (Q31431)}$>$ \\
\hline
(3) Is belgium bigger than MASK $<${\it new york (Q1384)}$>$ \\
(4) Why is MASK cooler than other places in Karnataka? $<${\it Bangalore (Q1355)}$>$ \\
\hline
(5) Whos faster john wall or MASK? $<${\it derrick rose (Q205326)}$>$ \\
(6) Who has more rings brian scalabrine or MASK? $<${\it  dirk nowitzki (Q44068)}$>$ \\
\hline
\end{tabular}
\caption{Comparative reasoning as fill-the-slot task: examples of entity-masked human-authored comparative questions in the target domains of {\it animals}, {\it cities/countries} and {\it NBA players}. The original masked span and its Wikidata entity identifier are given in brackets.}
\label{tab:examples}
\end{table}

In our experiments, we evaluate state-of-the-art transformer-based models~\cite{transformer} on this task. Specifically, we employ BERT~\cite{bert19}, which has been proven successful on a wide range of language understanding and knowledge modeling tasks~\cite{petroniEMNLP19,jiangTACL2020}. Notably, BERT represents a sentence as a sequence of sub-word terms, and can predict only a single term at a time. The evaluation of BERT on fill-the-blank fact completion tasks is therefore limited to the prediction of single-word entity mentions~\cite{petroniEMNLP19}. We thus also consider RELIC, an entity-focused variant of BERT~\cite{linkARXIV20}, which overcomes this limitation. The training procedure for RELIC involved predicting masked entity mentions in the whole of Wikipedia, where it has been shown that the resulting embeddings capture semantic entity properties~\cite{soaresLLD19,fevry2020}. We evaluate RELIC's ability to rank singe-term as well as multi-term entities as candidates in comparisons. We believe that this work is among the first efforts to perform sentence completion for multi-token entity mentions~\cite{fevry2020}.

Following common practice, we evaluate publicly available pretrained models on our task~\cite{petroniEMNLP19}, and also finetune the models on task-specific examples~\cite{richardsonAAAI2020}. While the results for the pretrained models are low, following in-domain finetuning, performance is close to human-level. However, more detailed analyses show that the performance gains achieved with finetuning are mainly due to lexical adaptation of the models, and do \emph{not} indicate any general-purpose modeling of comparative reasoning.
In fact, performance drops dramatically (down to levels seen prior to finetuning) when evaluated on out-of-domain examples--i.e., the models fail to generalize across domains. Overall, our analyses indicate that pretrained models can not perform the reasoning required to complete comparison questions, and further, that they fail to learn this skill from standard finetuning.

\section{Related Work}

In several recent works, researchers designed probing tasks and datasets so as to evaluate the extent to which large probabilistic language models capture world knowledge~\cite{petroniEMNLP19,talmorARXIV19,robertsARXIV20,richardsonAAAI2020,zhangEMNLP20}, as opposed to memorizing word co-occurrences and syntactic patterns~\cite{mccoyACL19}. Petroni {\it el al.}~\citeyearpar{petroniEMNLP19} presented LAMA, a set of cloze (fill-the-blank) tasks, to test the factual and commonsense knowledge encoded in language models. Their tasks require the prediction of a masked word in sentences involving relational facts drawn from different sources, including T-REx, a subset of Wikidata triples~\cite{trex}, common sense relations between concepts from ConceptNet~\cite{conceptnet}, and SQUAD, a Wikipedia-based question answering benchmark~\cite{squad16}. They concluded that the BERT-large model captures relational knowledge comparable to that of a knowledge base extracted from the reference text corpus using a relation extractor and an oracle entity linker. A followup work~\cite{kassnerACL20} showed however that large LMs, including BERT-large, ELMo~\cite{elmo}, and Transformer-XL~\cite{transofrmerxl}, are easily distracted by elements like negation and misprimes, suggesting that these models do well due to memorizing of subject and filler co-occurrences. Talmor {\it et al}~\citeyearpar{talmorARXIV19} similarly observed that high-performing finetuned LMs like RoBERTa~\cite{roberta} degrade to low performance when small changes are made to the input. Recently, Zhang \emph{et al.}~\citeyearpar{zhangEMNLP20} showed that contextual language models hold some knowledge about the scale of object properties, but fall short on common sense scale understanding. 

Our work continues this line of research, probing the capacity of large LMs to reason about world knowledge in natural language, while teasing apart lexical knowledge from semantic reasoning. The task and datasets of comparison completion presented in this work are comprised of human-authored questions, as opposed to synthetic examples. We further complement the evaluation with human judgements, highlighting challenges and gaps in contextual reasoning about entities and their properties, as perceived by humans. Unlike previous works, we extend slot filling to multi-token entity mentions.

Several prior works attempted answering comparative questions using large LMs~\cite{bagherAAAI16,forbesACL17,yangACL18,talmorARXIV19}.
For example, Yang {\it et al}~\citeyearpar{yangACL18} extracted relative object comparisons from pretrained word embeddings. In order to elicit whether an elephant was larger than a tiger, they compared the projection of the word embeddings for the object pair (``elephant'' and ``tiger'') to the embeddings of hand-coded adjectives, which denote the two poles of the target dimension (``big'' and ``small''). Unlike this work, they did not generate questions, or evaluate if compared pairs were perceived as sensible by humans. Talmor {\it et al}~\citeyearpar{talmorARXIV19} employed large neural LMs for a similar task, predicting the slot-filler for sentences like ``\emph{the size of a airplane is MASK than the size of a house}'' as either ``larger'' or ``smaller''.  Talmor \textit{et al} only considered comparisons based on two attributes, size and age, and did not formulate or complete questions.

Unlike these works, we do not seek to extract commonsense relations from text. Rather, we focus on automatically generating sensible comparisons, which are perceived as meaningful and non-trivial for a human.

\section{Comparative Question Dataset}
\label{sec:dataset}

We turn to question answering (QA) websites as a source of comparative questions. The questions posted in these sites are of interest to humans, reflecting common-sense as well as world knowledge in various domains (see Table~\ref{tab:examples}).

We constructed datasets of entity-masked user questions for the purpose of this research, focusing on three domains: {\it animals}, {\it cities and countries},\footnote{We sometimes abbreviate \emph{cities and countries} as \emph{cities}.} and {\it NBA players}.
To identify comparative questions at scale and with high quality we employed several heuristics (see also~\cite{tandon2014acquiring}), aiming at high-precision extraction of comparative questions in these domains out of a large pool of user questions crawled from various QA sites.

To this end, we syntactically parsed each question and annotated it with entity mentions using Google Cloud NLP API\footnote{\url{https://cloud.google.com/natural-language/}}. As part of this annotation, entity mentions were associated with their Wikidata identifiers.\footnote{www.wikidata.org} We first retained questions that contained at least one comparative word--a word with a part-of-speech tag of JJR or RBR. Then, using Wikidata's categorization system, we selected only questions that included two or more entity mentions of the target domains, e.g., we kept questions that contained two or more mentions of entities known to be animals.

To better understand the characteristics of these questions, we extracted the text span pertaining to the compared property based on the dependency structure of the question. Concretely, the relation was defined as the comparative word and its modifiers. We then manually grouped the lexical variations of the same comparative relation, and following, all questions that map to the same triplet $\{e_l,rel,e_r\}$ consisting of two compared entities $e_l,e_r$ and a relation $rel$. For example, ``\emph{Does lion run faster than a tiger}'' and ``\emph{who is faster tiger or lion}'' were both mapped to the same triplet $\{$\emph{lion, speed, tiger}$\}$. 

\begin{table}[t]
\small
\begin{tabular}{p{0.46\textwidth}}
\hline
\textbf{Animals:} intelligent, cute, clean, strong, speed, price, smelly, dangerous, size, life span, friendly, hearing, swimming speed, mass, length, height, deadly, venomous \\
\textbf{Cities, countries:} hot, cheap, safe, better, powerful, rich, large, popular, liberal, fun, beautiful, old, developed, populated, democratic, friendly, rainy, strong economically, culturally diverse, elevated, clean, better technologically, has more jobs, livable \\
\textbf{NBA players:} better, (better) shooting, career, clutch, defence, dunking, strength, popular, height, important, better season \\
\hline
\end{tabular}
\caption{Common compared properties per domain, listed by popularity (descending)}
\label{tab:properties}
\end{table}

As detailed in Table~\ref{tab:properties}, we found that the properties compared are often subjective, e.g., asking which animal is \emph{cuter}, or more \emph{intelligent}, and likewise, what country or city is more \emph{fun}, or more \emph{beautiful}. Some of the properties are factual yet not directly measurable, e.g., comparing animals by {\it strength}, cities by {\it liberalism}, and NBA players by their \emph{career}. Only a subset of the questions are factual and strictly quantitative, e.g., animal \emph{speed}, city \emph{population}, and athlete \emph{height}. Nevertheless, we noticed that all properties we encountered restrict, to some degree, the scope of entity pairs that make the comparison sensible. 
We note that since properties can be refined at various levels, e.g., ``\emph{London warmer than NYC in mid February?}'', and because questions may be realized in directional form, e.g., \emph{"Why is the black rhino more endangered than the white rhino?"}, the comparative relational triplets, as extracted using our heuristics, are not generally transitive, nor a complete characterization of the questions' semantics. 

We next split the collected questions into train, development and test sets by randomly sampling each semantic triplet $\{e_l,rel,e_r\}$ into one of the sets, together with all its mapped questions. This way, we prevented triplet memorization: exposing a model on one surface-form variation of a triplet at training time, while testing on a different variation. Finally, we chose one of the compared entities as the target prediction and masked it out.

The statistics of the split sets is presented in Table~\ref{tab:stats}. The examples sampled for testing were validated by human raters. Roughly 90\% of the sampled examples were verified as relevant comparative
questions, indicating that our rule-based approach for extracting comparative questions is highly precise. Irrelevant questions were removed from the test sets. Due to legal restrictions, we limited the source of the questions in the test sets to be only from \url{yahoo.answers.com}, and all test questions were manually screened to remove unethical questions. In our experiments, we confirmed that these test sets are representative, yielding results that are similar and consistent with test samples extracted randomly from multiple question answering Websites. Our test sets for the three domains are available at \url{https://github.com/google-research-datasets/comparative-question-completion}. To the best of our knowledge, this is the first publicly available collection of human-authored, and human-validated, natural comparative questions.

\begin{table}[t]
\small
\begin{tabular*}{\columnwidth}{llccc}
\hline
 &  & Animals & Cities & NBA \\
\hline
Test set & size & 1101 & 1261 & 1035 \\
& single-token slots & 62.1\% & 73.6\% & 3.5\% \\ 
& distinct masked entities & 521 & 798 & 375 \\
 & Human eval. subset & 200 & 229 & 145\\
 \hline
\multicolumn{2}{l}{In-domain train set size} & 31K & 37K & 7K \\
\multicolumn{2}{l}{Diverse train set size} & 1.15M & 1.05M & 1.2M \\
\hline
\end{tabular*}
\caption{Data statistics}
\label{tab:stats}
\end{table}

\section{Prediction Models}
\label{sec:methods}

We formulate comparison generation as an entity prediction task: given a comparative question we mask one of the compared entities' span, and ask a tested model to predict an entity (or term) in the masked entry.

Following the work of Petroni {\it et al.}~\citeyearpar{petroniEMNLP19}, which showed excellent performance using BERT-large~\cite{bert19} on factual slot-filling questions, we used BERT-large as one model for completing comparative questions.  BERT-large is a Transformer model that consists of 24 layers, 1024 nodes per layer, and 16 heads.
BERT scores its term vocabulary for each masked entry, and we take the resulting ranked list as its prediction. Since BERT only predicts single terms, it can only recover the original masked span for a subset of the questions. Yet, it may predict other single-term slot fillers that are perceived as meaningful.

Many human generated questions include multi-term entities as their comparands (see Table~\ref{tab:examples}). To measure the performance of neural LMs on multi-term entities, we chose RELIC~\cite{soaresLLD19,linkARXIV20} as our second tested model. RELIC learns vector representation of entities from the contexts in which those entities are mentioned. It uses BERT to encode texts in which a span known to denote an entity mention has been blanked out, training an entity encoder to match the BERT representation of the entity’s contexts. The compatibility between an encoded entity $e$ and the context $x$ is assessed as the scaled cosine similarity of their vector representations.

Our implementation of RELIC uses the pretrained BERT-small model, comprised of 12 layers, 768 nodes per layer, and 12 heads. The size of the entity embeddings in our experiments is $300$. We note that unlike BERT-large, which makes predictions out of a general English term vocabulary ($\sim$30K subwords), RELIC ranks all entities in Wikipedia.\footnote{As in prior work, we only consider the one million most popular entities in Wikipedia~\cite{fevry2020}.}

Prior work showed that the resulting entity representations learned by RELIC encode indicative information about the entities' semantic types and properties, as verified  through a few-shot Wikipedia category prediction task \cite{soaresLLD19}.  Also, the model correctly answered a majority of the questions on the entity-centric TriviaQA question-answering task~\cite{linkARXIV20,robertsARXIV20}. RELIC is therefore assumed to capture a significant amount of world knowledge about entities and their properties. 

\paragraph{Finetuning}

\begin{figure*}[t]
\centering
\includegraphics[width=5.72cm]{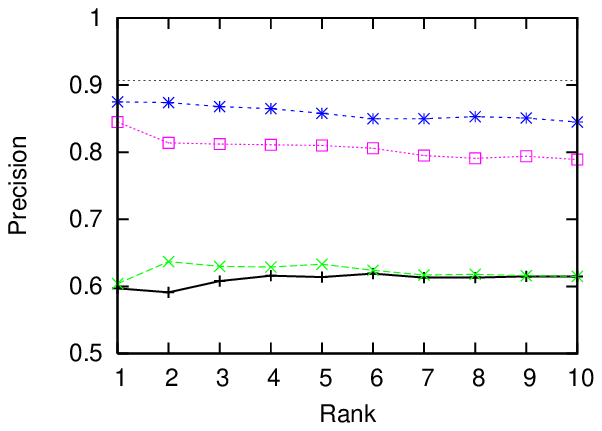}
\includegraphics[width=5.72cm]{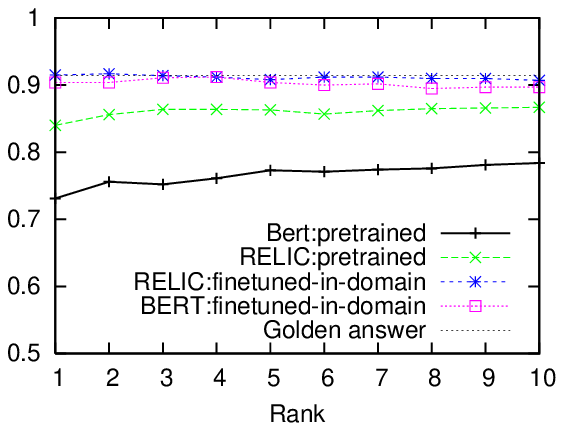}
\includegraphics[width=5.72cm]{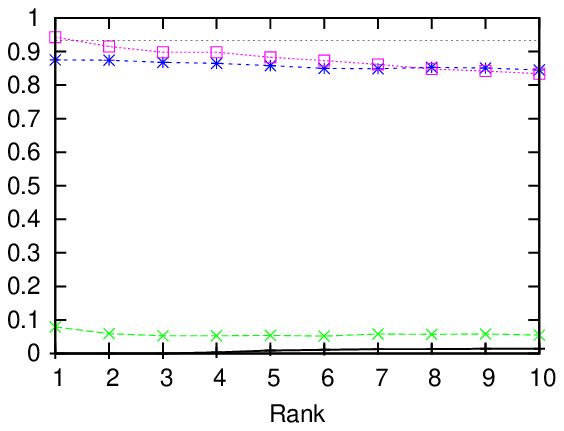}
\caption{Human evaluation: precision-at-rank 1--10 for Animals (left), Cities \& Countries (center) and NBA players (right).}
\label{fig:human}
\end{figure*}

Notably, neither BERT or RELIC have been trained on user-authored questions in general, and comparative questions in particular. The BERT-large model was pretrained using the BooksCorpus (800M words) (Zhu et al., 2015) and English Wikipedia (2,500M words), and RELIC was pretrained using hyperlinked entity mentions in the whole of English Wikipedia. 

We therefore apply the common practice of finetuning the tested models, so as to adapt them to the genre and task, using our large in-domain training datasets of comparative user questions (see Table~\ref{tab:stats}). We followed standard training procedure for finetuning, setting the number of epochs (varying between 1-5 in all cases) by optimizing performance on a held-out validation set, using an AdamW optimizer with a learning with rate of 2e-5. This resulted in six specialized LMs, one for each domain and method combination.

We ask: to what extent do these models improve on our task of comparison completion, having been finetuned using in-domain examples, and what do they learn via finetuning?

\paragraph{Evaluation}

Following the literature on question answering and knowledge-base completion, we evaluate each model based on how highly it ranks the masked ground-truth entity (RELIC) or token (BERT), measured by the metrics of Mean Reciprocal Rank (MRR) and Recall-at-1 (R@1)~\cite{baeza1999modern,nickel2015review}.
In general, however, there may be other entities, or terms, that are sensible in the context of a given comparative question. We therefore report the results of a complementary human evaluation, providing comprehensive assessment of the top-ranked responses of each model for a subset of the questions (Table~\ref{tab:stats}).\footnote{We sampled examples corresponding to distinct relational triplets in the test set of each domain.} In this evaluation mode, human raters were presented with the masked question, along with a predicted response, and were asked to assess whether the slot-filling candidate was sensible or not in the given context. 

Three raters evaluated each candidate, and the candidates were assigned the majority label. Inter-annotator agreement rates, assessed in terms of Fleiss Kappa, are \emph{moderate} (0.57) on \emph{animal} examples, \emph{substantial} (0.63) on \emph{cities and countries} examples, and \emph{nearly-perfect} (0.88) on \emph{NBA players} examples. These agreement rates indicate that humans hold a strong sense of what is perceived as sensible vs. non-sensible comparisons. Importantly, as we show below, the results of the human evaluation correlate with the MRR and R@1 metrics, validating the use of these metrics on our task.      

\section{Results}

\subsection{Pretrained Models}

Prior work showed that BERT and similar large models that are pretrained on large amounts of text, perform well on fill-in-the-blank tasks that require world knowledge, such as question answering \cite{petroniEMNLP19}. Accordingly, we evaluate pretrained BERT and RELIC on our task of auto-completing entity-masked comparative questions. 

The results are detailed in Table~\ref{tab:pretrained}. While the models' performances are shown side by side, we stress that BERT and RELIC are not directly comparable.
First, RELIC's vocabulary includes roughly 1 million entities, whereas BERT attends a much smaller vocabulary of $\sim$30K tokens. In addition, BERT cannot predict entity mentions that comprise more than a single (sub)word  (R\@1 and MRR are set to zero using BERT in multi-token entity prediction examples). As a more direct comparison on the prediction space, we also report in Table~\ref{tab:pretrained} the models' performance only on the subset of examples that mask out single-token entity mentions. These are the majority within the test questions about \emph{animals} (62\%) and \emph{cities} (73\%), but only a small fraction (3.5\%) of \emph{NBA players} examples (see Table~\ref{tab:stats}). 

\begin{table}[t]
\small
\begin{tabular*}{\columnwidth}{lcccccc}
\hline

 &  \multicolumn{2}{c}{Animals (A)} & \multicolumn{2}{c}{Cities (C)} & \multicolumn{2}{c}{NBA (N)} \\
  & R@1 & MRR &  R@1 & MRR &  R@1 & MRR \\
\hline
 &  \multicolumn{6}{l}{FULL SET:} \\
RELIC & 0.120 & 0.193 & 0.102 & 0.210 & \textbf{0.008} & \textbf{0.019} \\
BERT & \textbf{0.182} & \textbf{0.240} & \textbf{0.136} & \textbf{0.229} & 0.004 & 0.005 \\
\hline
 &  \multicolumn{6}{l}{SINGLE-TOKEN ENTITIES:} \\
RELIC & 0.184 & 0.290 & 0.129 &	0.257 & 0.056 &	0.095 \\
BERT & \textbf{0.292} & \textbf{0.386} &
\textbf{0.292} & \textbf{0.386} & \textbf{0.111} & \textbf{0.144} \\
\hline
\end{tabular*}
\caption{Results using the pretrained models}
\label{tab:pretrained}
\end{table}

The figures in Table~\ref{tab:pretrained} are generally low and indicate that large pretrained language models like BERT and RELIC often fail in making sensible comparisons. Such low figures are also reported on related factual slot filling tasks. For example, slot-filling performance of BERT-large on multi-answer (N-to-M) relational facts measured 0.24 in R\@1 on ~\cite{petroniEMNLP19} vs. R\@1 of 0.11--0.29 across domains in our experiments. 
As shown in the table, pretrained BERT significantly outperforms pretrained RELIC when evaluated on masked single-token prediction. On the full test-sets BERT still outperforms RELIC on \emph{animals} and \emph{cities}, in which the majority of gold predictions are single tokens, but not on \emph{NBA players}, which is also the domain in which both models perform the worst. We also observe that RELIC performs better on a subset of single-token entities. One possible reason is that single word entity mentions are more frequent in Wikipedia compared to multi-word entities, and are therefore better represented in pretrained RELIC.

\begin{table}[t]
\small
\begin{tabular}{p{0.02\textwidth} p{0.01\textwidth}p{0.38\textwidth}}
\hline
 & \multicolumn{2}{l}{\textbf{Pretrained:}} \\
(1) & R &  Dragon, Lion, Pig, Crocodile, Chicken \\
 & B & lion, bear, tiger, wolf, dog \\
(2) & R & Columbidae, Sparrow, Goose, Rabbit, Dog \\
& B & dog, bird, mouse, beetle, butterfly \\
(3) & R & France, Belgium, Germany, Denmark, Flanders \\
& B & france, germany, canada, england, russia \\ 
(4) & R & Bangalore, Mysore, Visibility, Chennai,  Humidity \\
& B & mysore, bangalore, it, karnataka, this \\
(5) & R & Dome, Ceiling, Pinnacle, Roof, Truss \\
 & B & me, what, you, i, who \\
(6) & R & Thumb, Wiley (musician), Sasha (DJ), Milo (musician), Texas (band) \\
    & B & me, tiffany, himself, him, you  \\
\hline
 & \multicolumn{2}{l}{\textbf{Finetuned:}} \\    
  
(1) & R & American black bear, Cougar, Polar bear, Snow leopard, Canada lynx \\
 & B &  wolf, coyote, lion, leopard, tiger \\
 
 (2) & R & Parrot, Finch, Budgerigar, Dog, Columbidae \\
& B & parrot, dove, bunny, duck, dog \\

(3) & R & Netherlands, United States, United Kingdom, Czech Republic, New Zealand \\
& B & france, canada, germany, spain, netherlands \\

(4) & R & Mangalore, Bangalore, Mysore, `Hassan, Karnataka', Bellary \\
& B & bangalore, mysore, hassan, hyderabad, chennai \\
(5) & R & Derrick Rose, Brandon Jennings, Jeremy Lin, Brook Lopez, Andre Drummond \\
 & B &  kobe, curry, wade, harden, devin \\
 
(6) & R & Dwyane Wade, Scottie Pippen, Draymond Green, Dirk Nowitzki, Tracy McGrady \\
    & B & kobe, wade, curry, jordan, jamal \\
\hline
\end{tabular}
\caption{Top-ranked predictions using the pretuned and finetuned versions of BERT (B) and RELIC (R)}
\label{tab:top}
\end{table}

We further analyzed these results to gauge the extent to which auto-completed questions reflect common-sense reasoning and world knowledge, producing \emph{sensible} as opposed to nonsensical comparative questions.
Table~\ref{tab:top} lists the five top-ranked predicted slot-filler per model for the example questions listed in Table~\ref{tab:examples}. These cases exemplify various failures in capturing comparison semantics. First, there are mismatches in the expected entity type, e.g. RELIC suggesting ``thumb'', ``dome'' and ``ceiling'' instead of NBA players.
Second, BERT often ranks high common words, such as ``it'', ``himself'' and ``me''. Third, the models fail to make common-sense inferences, resulting in some nonsensical questions. For example, RELIC offers to compare the size of a ``cougar'' with a ``dragon'' or a ``chicken'' (ex.1). In another example, when comparing ``loyalty'' for a pigeon, BERT assigns high scores to the nonsensical slot-fillers of ``bird'' and ``beetle'' (ex.2). Also, BERT suggests comparing the size of the rather small Belgium with very large countries like Canada and Russia (ex.3).

Our human evaluation of the pretrained models (Figure~\ref{fig:human}) reflects these observations. The ratio of sensible answers among the top-ranked responses by BERT is roughly 0.60 for \emph{animals}, 0.73 for \emph{cities} and practically zero for \emph{NBA players}. The responses by RELIC were perceived as slightly more sensible by the raters, with sensible answer ratio of 0.60 (\emph{animals}), 0.84 (\emph{cities}) and 0.08 (\emph{NBA players}). We note that the human evaluation was rather permissive. For example, all countries predicted by RELIC and BERT for the question \emph{"Is belgium bigger than MASK?"} were labeled as sensible, except for `Belgium' and `Flanders' (Table~\ref{tab:top}).

\subsection{Finetuned Models}

There are notable differences between the texts on which BERT and RELIC were pretrained and our test data, in terms of genre, data distribution, and semantics. A standard approach to close this gap is to finetune pretrained models on training examples that are similar to those in the test-set. But, does such finetuning really learn generic capabilities for solving comparative question completion?

To answer this question, we finetuned the tested models using our in-domain training sets of comparative questions, initializing the models' parameters to those of the pretrained models. For each method, we applied its default training procedure, i.e., a fixed ratio (15\%) of random token masking for BERT,\footnote{As our examples consist of individual sentences, we duplicated the sentences, yet ignored the next-sentence prediction component.} and entity masking for RELIC. This resulted in specialized models per domain. 

\begin{table}[t]
\small
\begin{tabular*}{\columnwidth}{lcccccc}
\hline
 &  \multicolumn{2}{c}{Animals (A)} & \multicolumn{2}{c}{Cities (C)} & \multicolumn{2}{c}{NBA (N)} \\
  & R@1 & MRR &  R@1 & MRR &  R@1 & MRR \\
\hline
RELIC (R) & \textbf{0.540} & \textbf{0.640} & \textbf{0.358} & \textbf{0.481} & \textbf{0.439} & \textbf{0.545} \\
BERT (B) & 0.308 & 0.370 & 0.190 & 0.286 & 0.047 & 0.050 \\
\hline
Co-occur. & 0.327 & 0.447 & 0.255 & 0.374 & 0.367 &	0.478 \\
\hline
R:resample & 0.262 & 0.390 & 0.258 & 0.388 & 0.177 &	0.276 \\
B:resample & 0.190 & 0.264 & 0.134 &	0.233 & 0.020 &	0.026 \\
\hline
\end{tabular*}
\caption{In-domain finetuning results and baselines}
\label{tab:finetuned}
\end{table}

Table~\ref{tab:finetuned} details the results of the finetuned models, showing a dramatic increase in performance. RELIC reached R@1 of 0.54 and MRR of 0.64 on \emph{animals}, improving by 157\% and 232\% respectively over the pretrained model. Similar improvements are observed for \emph{cities}, reaching R@1 of 0.36 and MRR of 0.48 (relative improvement of 250\% and 128\%). In the domain of \emph{NBA players}, for which the pretrained model results were poor, similar excellent performance was obtained with finetuning, reaching 0.44 and 0.55 in R@1 and MRR.   

While the finetuned BERT models also present large improvements over pretrained BERT, finetuned RELIC significantly outperforms finetuned BERT on all domains. Crucially, RELIC has globally improved on entity slot-filling prediction, whereas BERT improvements remained limited to single-token examples. 
Figure~\ref{fig:human} shows similar trends based on the human evaluation, where both finetuned RELIC and BERT models yield predictions that are mostly perceived as meaningful: the ratio of sensible answers is 0.85-0.92 among the top 10 predictions across domains for finetuned RELIC, and 0.79-0.94 for finetuned BERT. These results are almost as high as the human judgement scores of the original masked entity. 

Finally, Table~\ref{tab:top} also lists the top predictions produced by the finetuned models for the examples in Table~\ref{tab:examples}. Clearly, both BERT and RELIC improved by predicting slot-fillers of the expected semantic type. This is especially apparent in the \emph{NBA players} domain, where predictions now include player names. The finetuned models failed however on questions such as \emph{"who was a better sixer and had a better career, charles barkley or MASK?"}---the top-scoring predicted slot-fillers included NBA players who never played for the `Philadelphia 76er' team. 

\paragraph{Memorizing vs. learning.}
The positive impact of LM finetuning has been previously demonstrated for a variety of tasks~\cite{peters2020,richardsonAAAI2020,zhaoACL2020}. Next we ask: does the success of finetuning RELIC and BERT in our experiments imply that these models have learned comparison semantics? or, are the observed performance gains merely due to memorizing of relevant lexical statistics?

\begin{figure}[t]
\centering
\includegraphics[width=0.9\columnwidth]{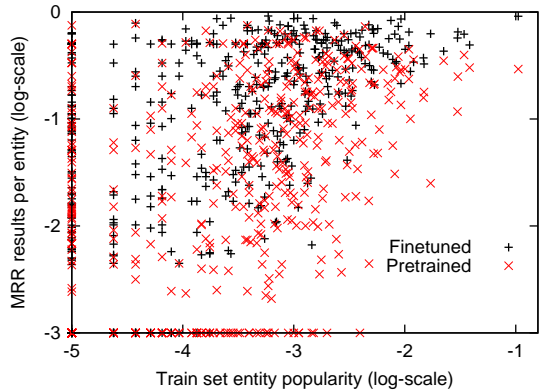}
\small
\caption{Fill-in-the-blank \emph{cities} test results using pretrained and finetuned RELIC: MRR vs. entity popularity (frequency among all of the masked entity mentions in the train set) (log-log scale). The top-right point stands for the popular slot filler of `United States'.}
\label{fig:freq}
\end{figure}

Figure~\ref{fig:freq} offers a visualization of the statistical information that is encoded and learned by pretrained and finetuned RELIC.
The figure depicts average MRR results for each of the masked entities in the \emph{cities} test set, contrasted with their popularity (relative frequency) within the training set. As shown, while performance using the pretrained RELIC is weakly correlated with train set entity frequency, strong correlation is observed following finetuning; for example, nearly perfect MRR is obtained on questions where the masked entity is ``United States'' -- the most frequent entity in the train set. Significantly lower levels of prediction success are obtained for low-frequency masked entity types. 

We further explored the statistics that might have been learned from our training sets during finetuning. Based on the annotation of entity mentions, we considered a baseline that scores candidate entities by their co-occurrence frequency with the compared entity within the train set. According to this context-less baseline, given the question \emph{"which one is cuter MASK or cats?"}, the top-ranking answer is dogs, as ``dogs'' are the entity that most frequently co-occurs with ``cats'' in the training examples.
Table~\ref{tab:finetuned} presents these results, showing that the entity co-occurrence baseline (`co-occur.') yields strong results, outperforming the pretrained models, and explaining some of the performance gains obtained by the finetuned entity-focused RELIC. Still, RELIC outperforms this baseline by a substantial margin. Part of this lies in the linguistic and word knowledge learned by RELIC during the pretraining phase, indicating that deep entity representation generalizes beyond discrete lexical co-occurrences information.

Finally, to tease apart the learning of corpus statistics versus semantics, we conducted another experiment, in which we manipulated the train set distribution. Specifically, we reconstructed the training datasets by sub-sampling questions pertaining to frequent entity pairs, and over-sampling questions about infrequent entity pairs. Consequently, while we have not modified individual comparative questions in the train set, we have eliminated global information about the relative popularity of slot-fillers given each context entity. We then finetuned RELIC and BERT using the resampled train sets. The results are detailed in Table~\ref{tab:finetuned}, showing that performance for both RELIC and BERT plummets as a result of re-sampling. We therefore conclude that much of the performance gains following finetuning are in fact due to the memorizing of entity co-occurrence information, as opposed to learning comparative semantics.

\paragraph{Failure to generalize.}

\begin{table}[t]
\small
\begin{tabular*}{\columnwidth}{l|ccc|ccc}
\emph{Test} $\rightarrow$  &  \multicolumn{3}{c|}{BERT} & \multicolumn{3}{c}{RELIC} \\
  \hline
\emph{Train} $\downarrow$ & A & C & N & A & C & N \\
\hline
Animals & \cellcolor{gray!25}0.370 & 0.184 & 0.005 & \cellcolor{gray!25}0.640 & 0.205 & 0.009 \\
Cities & 0.237 & \cellcolor{gray!25}0.286 & 0.006 & 0.204 & \cellcolor{gray!25}0.481 & 0.030 \\
NBA & 0.172 & 0.153 & \cellcolor{gray!25}0.050 & 0.126 & 0.153 & \cellcolor{gray!25}0.545 \\
\hline
\end{tabular*}
\caption{Cross-domain fill-in-the-blank finetuning results}
\label{tab:cross}
\end{table}

\begin{table}[t]
\small
\begin{tabular*}{\columnwidth}{lcccccc}
\hline
 &  \multicolumn{2}{c}{Animals (A)} & \multicolumn{2}{c}{Cities (C)} & \multicolumn{2}{c}{NBA (N)} \\
  & R@1 & MRR &  R@1 & MRR &  R@1 & MRR \\
\hline
RELIC & 0.030 & 0.098 & 0.107 & 0.219 & 0.007 & 	0.018 \\
BERT & 0.153 & 0.216 & 0.132 & 0.229 & 0.007 & 0.010 \\
\hline
\end{tabular*}
\caption{Fill-in-the-blank results following finetuning using more than 1M topic-diverse comparative questions}
\label{tab:large}
\end{table}

If finetuned models have learned general comparison semantics, they should also improve on out-of-domain comparative questions. We tested the generalization capacity of the finetuned models through cross-domain evaluation, training on one domain and testing on the other two domains in our dataset. Table~\ref{tab:cross} depicts the results of this cross-domain evaluation. As shown, the boost in performance within-domain (the highlighted diagonal cells) does not carry to other domains. In fact, cross-domain performance for both BERT and RELIC drops back to pretrained performance levels, and sometimes below that. For example, RELIC performance on \emph{cities} questions rose from 0.210 to 0.481 in MRR following finetuning on in-domain \emph{cities} examples, whereas performance on \emph{cities} is as as low 0.184 and 0.153 with RELIC being finetuned on the out-of-domain examples of \emph{animals} and \emph{NBA players}, respectively.

Presumably, finetuning using examples sampled from a data distribution that is limited to a specific subject-domain might bias the learning process, adapting the model by learning relevant information at lexical level, as opposed to deeper, and more general, semantic structures. To address this concern, and encourage the learning of comparative semantics, we constructed a large and diverse training corpus of comparative questions, as follows. 

In our domain-specific datasets, all questions mention at least two entities of the target-domain semantic type, e.g., \emph{animals}.
Here, we rather considered all questions that contained some comparative word (JJR/RWR) and also contained one of the words: `than', `between' and `or'. This yielded a large corpus of over 1.3 million questions. We manually assessed a sample of these questions, and found roughly half of them to be comparative factual questions. We note that the inclusion of non-comparative questions should not hurt the learning process considering that the
models have been pre-trained on general texts. Finally, to minimize overlap between the train and test examples~\cite{lewis2020question}, we made 3 copies of the training examples, one per tested domain, and eliminated from each copy any question that included an entity mentioned or masked in the test questions of the respective target domain. 

We performed finetuning for all models, this time with the substantially larger and diverse training sets. The number of learning epochs was set using in-domain development examples. In all cases, a single training epoch provided the best results on the development sets. Table~\ref{tab:large} shows our evaluation results. We observe that finetuning using this large training set gives comparable and sometimes lower performance with respect to the pretrained models. From our cross-domain experiments as well as the large but topically non-overlapping training experiments we conclude that finetuning BERT and RELIC, state-of-the-art large language models, on complex semantic and linguistic structures such as comparisons, fails to generalize beyond the lexical level.

\section{Conclusions}

In this work, we introduced the task of predicting masked entity mentions in human-generated comparative questions. This task requires world knowledge, as well as language understanding and semantic reasoning. We have shown that finetuning BERT, and RELIC, a related entity-focused language model, on collections of comparative questions yields high performance on within-domain questions. However, we found that predictions are often perceived as nonsensical, and that learning by finetuning is limited to lexical phenomena, failing to generalize to out-of-the-domain examples. 

Tasks such as understanding and making comparisons can be used to probe whether enhanced models are capable of capturing and reasoning about entities and their properties. We publish our test-sets of human-authored comparative questions in the hope that they will serve as a benchmark for measuring complex understanding and reasoning in such models at standards of human interaction. 

Finally, considering that excellent performance is achieved on comparison completion using large language models finetuned on in-domain examples, we think that a similar framework can be used to create various sensible comparative questions in concrete subject domains. Such questions can be integrated by an agent to enhance the quality of dialogues and other types of interactions with humans.

\begin{small}
\bibliography{main}
\end{small}

\end{document}